\documentclass{article}
\usepackage[margin=1.25in]{geometry}
\usepackage[utf8]{inputenc}
\usepackage{graphicx}
\usepackage{subfig}
\usepackage{multirow}
\usepackage[square,numbers]{natbib}
\usepackage[english]{babel}
\usepackage{multicol}
\usepackage[affil-it]{authblk}
\title{Not cool, calm or collected: Using emotional language to detect COVID-19 misinformation}
\author[1,*]{Gabriel Asher}
\author[1]{Phil Bohlman}
\author[1]{Karsten Kleyensteuber}

\affil[1]{Dartmouth College, Computer Science}
\affil[*]{Corresponding author: gabriel.l.asher.24@dartmouth.edu}
\date{February 2023}

\begin{document}

\maketitle

\section{Abstract}

COVID-19 misinformation on social media platforms such as twitter is a threat to effective pandemic management. Prior works on tweet COVID-19 misinformation negates the role of semantic features common to twitter such as charged emotions. Thus, we present a novel COVID-19 misinformation model, which uses both a tweet emotion encoder and COVID-19  misinformation encoder to predict whether a tweet contains COVID-19 misinformation. Our emotion encoder was fine-tuned on a novel annotated dataset and our COVID-19 misinformation encoder was fine-tuned on a subset of the COVID-HeRA dataset. Experimental results show superior results using the combination of emotion and misinformation encoders as opposed to a misinformation classifier alone. Furthermore, extensive result analysis was conducted, highlighting low quality labels and mismatched label distributions as key limitations to our study. 

\section{Introduction}
COVID-19 is a worldwide pandemic which, as of December 2022, has killed over 14 million people \cite{msemburi2022estimates}. Exasperating mortality are swaths of COVID-19 misinformation propagated on social media platforms such as Twitter, Facebook, Youtube, and Instagram \cite{cinelli2020covid}. Much of misinformation is related to topics such as vaccine hesitancy, virus origins, virus transmissions, and unsubstantiated medical advice \cite{nuzhath2020covid, brennen2020types}. These misinformation campaigns can have deadly impacts. COVID-19 vaccines, which have been clinically cleared by drug regulatory agencies such as the FDA, have proven to be effective for up to 95\% of patients \cite{mahase2020covid, meo2021covid}. This vaccine protection is especially important for high-risk demographics such as the elderly and immuno-compromised. Thus, vaccine hesitancy and misinformation online could have adverse effects on the number of recipients of the vaccine, potentially leading to otherwise preventable deaths.

Several efforts have been made to apply machine learning advances to COVID-19 misinformation detection. Shu et al. created dEFENSE, a generalized misinformation detection system which uses separate word, sentence, and comment encoders \cite{shu2019defend}, which has been applied to COVID misinformation datasets \cite{alenezi2021machine}. Several research studies have used BERT fine-tuning approaches to detect misinformation \cite{hossain2020covidlies, alenezi2021machine, dharawat2022drink, pranesh2021looking}. Additionally, several baseline datasets have been created. Patwa et al. have a 10700 post, manually annotated COVID-19 misinformation dataset from 2021 \cite{patwa2021fighting}. Cui and Lee created CoAID, a repository of over 200000 annotated tweets, articles, and social platform posts \cite{cui2020coaid}. Kim et al. create a dataset of 722 COVID-19 claims, which they use to create over 150000 annotated tweets \cite{kim2021fibvid}. Dharawat et al. created a dataset of 61286 tweets, manually labeling the degree of COVID-19 misinformation \cite{dharawat2022drink}.

Emotions play a significant role in the spread of misinformation, as people are more likely to believe and share information that confirms their existing beliefs and fears. For example, people who are anxious about the pandemic may be more likely to believe false information about a supposed cure or treatment, as it offers them a sense of comfort and hope. On the other hand, those who are skeptical of government actions may be more likely to believe conspiracy theories that downplay the severity of the virus. These emotions may lead to the rapid spread of false information, causing confusion, fear, and harm to public health efforts to contain the pandemic.

Encoding external data in addition to task-specific knowledge has showed success in other domains of natural language processing. Lokala et al. train a symptom classifier for cardiovascular disease (CVD) using separate external (gender) and task-specific (symptomatic) encoders \cite{lokala2022computational}. They observe better results using both of these encoders as opposed to a task-specific encoder alone. Furthermore, Twitter has a reputation for toxic, and therefore heavily emotive, language \cite{norton_2022}. These emotions are especially pronounced in regards to tweets associated to COVID-19 outbreaks \cite{manguri2020twitter}. This leads to our main research question: to what extent does encoding tweet emotions into COVID-19 misinformation detection models improve model accuracy? This primary research question leads to two secondary questions. What is the relationship between tweet emotion and COVID-19 misinformation? What are the trends in correctly and incorrectly predicted misinformation labels?

Based on these promising approaches to using external data to assist predictions in mental-health datasets, we believe that similar strategies could be applied to COVID-19 misinformation data. Both datasets contain charged language, and because of the short length of tweets, strong words are used more than in other social media sites. Thus, we create \textit{emotion}, the first COVID-19 misinformation system which leverages emotional encoding in addition to contextual and factual encoding to improve predictive accuracy.

\section{Methods}

\subsection{Dataset and Data Aquisition}
\subsubsection{COVID Misinformation}
Our dataset is a subset of the COVID-HeRA dataset created by Dharawat et al. \cite{dharawat2022drink}, This dataset maps tweets to 5 degrees of misinformation: real news/claims, possible severe, highly severe, other, and refutes/rebuts misinformation. These labels were annotated manually by two fluent English speakers, and Cohen's kappa coefficient between the annotators was 0.7037, which shows strong alignment amongst ground truth labels \cite{dharawat2022drink}. The dataset consists of 61,286 tweets across these 5 classes (Table \ref{dataset breakdown}). However, given computational constraints and data imbalance issues, we filtered down our dataset to a total of 6,355 tweets. 

\begin{table}[!h]
\begin{center}
\begin{tabular}{| c | c || c | c | c |} 
 \hline \hline

 Category & \#Tweets COVID-HeRA & \#Tweets (ours) & Training Set & Test Set \\ [0.5ex] 
 \hline
 Possibly severe & 439 & 277 & 172 & 105\\ 
 Highly severe & 568 & 337 & 312 & 25\\
 Refutes/Rebuts & 447 & 350 & 89 & 261\\
 Other & 1,851 & 1,391 & 731 & 660\\
 Real News/Claims & 57,981 & 4,000 & 3,033 & 967\\ [1ex] 
 \hline
 Total & 61,286 & 6,355 & 4,337 & 2,018\\
 \hline \hline
\end{tabular}
\caption{\label{dataset breakdown} Dataset Statistics}
\end{center}
\end{table}

\subsubsection{Emotion Dataset}
From our COVID-HeRA dataset, we create an additional dataset which maps tweets to the following emotions: anger, disgust, fear, joy, sadness, surprise and neutral. This dataset is manually annotated by three native English-speaking annotators as follows. We randomly selected 100 unique tweets from our training set. Our labelers assign one of the 7 emotions to each tweet (Table \ref{emotion_data_breakdown}). Following this step, we use a transfer learning approach to fine-tune a Sentence BERT (SBERT) base model using our training data \cite{reimers-2019-sentence-bert}. This fine-tuning step is performed with a 90/10 train/test split of our manually annotated data. The agreement between this fine-tuned model and our labels is 72\%. Furthermore, the kappa score of our annotators, a measure of inter-annotator agreement, is 0.54, representing moderate agreement. Examples of tweet emotion annotations can be seen in Table \ref{label examples}.

\begin{table}[]
    \begin{center}
    \begin{tabular}{|c|c|}
    \hline \hline
        Emotion & \#Tweets \\ [0.5ex] 
        \hline
        Anger & 10 \\
        Disgust & 1 \\
        Fear & 10 \\
        Joy & 1 \\
        Neutral & 76 \\
        Sadness & 2 \\
        Surprise & 0 \\
        \hline
        Total & 100\\
        \hline \hline
    \end{tabular}
    \caption{\label{emotion_data_breakdown} Annotated Tweet Data}
    \end{center}
\end{table}

\begin{table}[ht]
    \centering
    \begin{tabular}{|p{0.15\linewidth} | p{0.6\linewidth}|}
    \hline
    Emotion label & Source text \\
    \hline
    Anger & @LorenSethC @SethAbramson Remember the "Coronavirus Hoax" that was invented just to hurt the president? Wonder what happened with that one. Fox News was quick to sponsor disbelief in the science on that. \\
    \hline
    Disgust & LGBTQ Demanding to Be Treated FIRST for Coronavirus (Because They’re So Disease-Ridden Already) https://t.co/D1RswYgvhh \\
    \hline
    Fear & Deep State Patented Coronavirus the perfect storm BREAKING: Coronavirus Hits 15\% Fatality Rate, 83\% Infection Rate for Those Exposed; Lancet Publishes Early Study That Points to Alarming Consequences for Humanity https://t.co/deJT76g5wE @realDonaldTrump  More death\\
    \hline
    Joy & Please join me in showing appreciation for those working during to deliver our essential needs. Stand on your front porch and clap, ring bells or bang pots for law enforcement, first responders, healthcare workers, grocers and others. They're our true heroes. \#19Thanks https://t.co/W9pFnxClbx \\
    \hline
    Neutral & Trump Signs Into Law \$2 Trillion Coronavirus Relief Package https://t.co/fcudCtbl4P \\
    \hline
    Sadness & Today my daughter-in-law’s uncle died from Covid-19 after a 2 month fight. We thought he was getting better, but took a turn and died this morning. We are all heartbroken \newline \newline Please stay home, protect yourself and your loved ones.\\
    \hline
    Surprise & \textit{NA - None of the sampled tweets were labeled as "Surprise"} \\
    \hline
    \end{tabular}
    \caption{\label{label examples} Example labeled tweets in hand-annotated dataset}

\end{table}

\subsection{Proposed Methodology}

In order to answer our research question, we propose the following methodology, which consists of two components. Firstly, we train two separate, task-adaptive SBERT pre-trained encoders to encode our tweets. The first pre-trained encoder is trained to classify the dominant emotion of a given tweet, using our emotion dataset. The second pre-trained encoder is trained to determine the severity of COVID misinformation. Our model architecture leverages these two pre-trained encoders as follows. Firstly, for a given input tweet, we generate emotion and misinformation embeddings. We extract the CLS tokens from each of these embeddings, and concatenate them. Then, we feed this concatenated embedding into a multi-layer perceptron (MLP) to generate an output label.

\subsection{Experiments}
\subsubsection{Experimental Setup}

In preparing for our experiment, we perform additional processing steps. First we filter out tweets that were no longer available (due to deletion or removal). Next, we remove mentions and links from each tweet using simple regular expressions. Finally, we randomly divide our data into 80\% training and 20\% test partitions.

\subsubsection{Baseline}

To establish baseline performance with which to compare our emotion-encoded embeddings, we performed experiments with three different models. In each, we pass only the cleaned tweet into an encoder to generate embedding vectors. Our first baseline model encodes each token in a tweet with their term-frequency inverse document frequency (TFIDF), a numerical representation of token importance with relation to the document. Strings of token TFIDFs are used to form feature vectors. We then pass them to a Random Forest Classifier to generate the label. The Random Forest Classifier uses random feature selection to generate decision trees used for making an aggregate classification. Our next baseline takes the average GloVe embedding of each word in a given tweet and passes it to a Logistic Regression classifier to generate a classification. The GloVe model encodes tokens in a meaning space defined by the probability of word combinations co-occurring in a document \cite{pennington2014glove}. We use a version of GloVe that has been pre-trained on a corpus of 6 billion tokens (400,000 word vocabulary) from Wikipedia and the Gigaword5 dataset. In our final baseline model, we use SBERT on the entirety of each tweet to generate feature vectors \cite{reimers-2019-sentence-bert}, which we then pass into an MLP to generate output predictions. SBERT modifies BERT, the original bi-directional encoder model for natural language tasks, with siamese and triple networks structures to generate semantically meaningful embeddings for sentences as opposed to single tokens \cite{DBLP:journals/corr/abs-1810-04805}. Specifically, we use the "all-MiniLM-L6-v2" model, which has been pre-trained on 1 billion sentence pairs. We then pass these feature vectors into a Logistic Regression classifier to generate a label.

We evaluate each model's performance on a multi-class labeling task (how severe of misinformation does the tweet contain, if at all?). 

\section{Results}
We summarize the results of our experiments in Table \ref{summary results}. 

\begin{table}[!ht]
    \centering
    \begin{tabular}{|ll|c|c|c|c|c||c|c|c|}
    \hline
        \multicolumn{2}{|l|}{Experiment} & \shortstack{\\ Real News \\ / Claims} & \shortstack{Refutes \\ / Rebuts} & Other & \shortstack{Possibly \\ severe} & \shortstack{Highly \\ severe} & Accuracy & \shortstack{Macro \\ avg.} & \shortstack{Weighted \\ avg.} \\ \hline
          \multirow{4}{*}{Precision} 
          & TFIDF & \textbf{0.58} & 0.29 & 0.58 & 0.01 & 0.00 & - & 0.29 & 0.50 \\ 
          & GloVe & 0.55 & \textbf{0.47} & 0.65 & 0.52 & 0.02 & - & 0.44 & 0.56 \\ 
          & SBERT & 0.55 & 0.33 & 0.79 & 0.70 & 0.04 & - & 0.48 & 0.60 \\ 
        & Emotion & \textbf{0.58} & 0.29 & \textbf{0.82} & \textbf{0.83} & \textbf{0.11} & - & \textbf{0.54} & \textbf{0.64} \\ \hline
          \multirow{4}{*}{Recall} 
          & TFIDF & 0.54 & 0.01 & \textbf{0.54} & 0.05 & 0.00 & - & 0.23 & 0.44 \\ 
          & GloVe & 0.95 & 0.03 & 0.09 & 0.38 & \textbf{0.12} & - & 0.31 & 0.51 \\ 
          & SBERT & \textbf{0.98} & 0.09 & 0.20 & 0.27 & 0.04 & - & 0.32 & 0.56 \\ 
        & Emotion & \textbf{0.98} & \textbf{0.16} & 0.42 & \textbf{0.56} & 0.09 & - & \textbf{0.37} & \textbf{0.60} \\ \hline
        \multirow{4}{*}{F1 Score} 
          & TFIDF & 0.56 & 0.02 & \textbf{0.56} & 0.02 & 0.00 & 0.44 & 0.23 & 0.45 \\ 
          & GloVe & 0.69 & 0.06 & 0.16 & 0.44 & 0.04 & 0.51 & 0.28 & 0.42 \\ 
          & SBERT & 0.71 & 0.14 & 0.32 & 0.39 & 0.04 & 0.56 & 0.32 & 0.48 \\ 
        & Emotion & \textbf{0.73} & \textbf{0.16} & 0.42 & \textbf{0.56} & \textbf{0.09} & \textbf{0.60} & \textbf{0.39} & \textbf{0.54} \\ \hline
    \end{tabular}
    \caption{\label{summary results} Summary of Results}
\end{table}

\subsection{Discussion of Results}

We show improvement in the multi-class classification of misinformation with our architecture (Emotion) compared to the chosen baseline models (TFIDF, GloVe, SBERT). Our model is marginally more precise when identifying Real News/Claims, shows worse performance for tweets that Refute or Rebut misinformation, but shows marked improvement in the various categories of tweets that contain or assert misinformation: Other, Possibly severe, and Highly severe. A similar pattern can be observed in Recall and F1-Score: marginally improvement in tweets containing real news or misinformation rebuttals, and material performance improvements in the categories of misinformation. We believe that special attention should be paid to our model's recall, which we believe is the most important measure for assessing the its effectiveness in a "real world" commercial application, such as assisting with the moderation of Twitter. In such a scenario, the model might not be tasked with definitively determining if a tweet contains misinformation; rather, it would generate a pipeline of tweets for a human moderator to review. With this in mind, minimization of false negatives, or the maximization of recall, would be the most important measure of model performance. In recall, we show our largest relative and absolute performance increases relative to the baseline.

We also show improvements over the baseline on the classification task in aggregate, with our architecture generating the highest Macro and Weighted Average Precision, Recall, F1 Score and Accuracy relative the to the baselines.

Also of note is that our dual-SBERT encoder model architecture (one fine-tuned on COVID-19 misinformation, the other fine-tuned for emotion classification) shows consistently higher performance than the directly comparable single encoder SBERT baseline, suggesting that our dual-encoding approach is effective in improving performance in this task. 

\section{Discussion}
\subsection{Model Interpretability}

\begin{figure}%
    \centering
    \includegraphics[width=16cm, height=8cm]{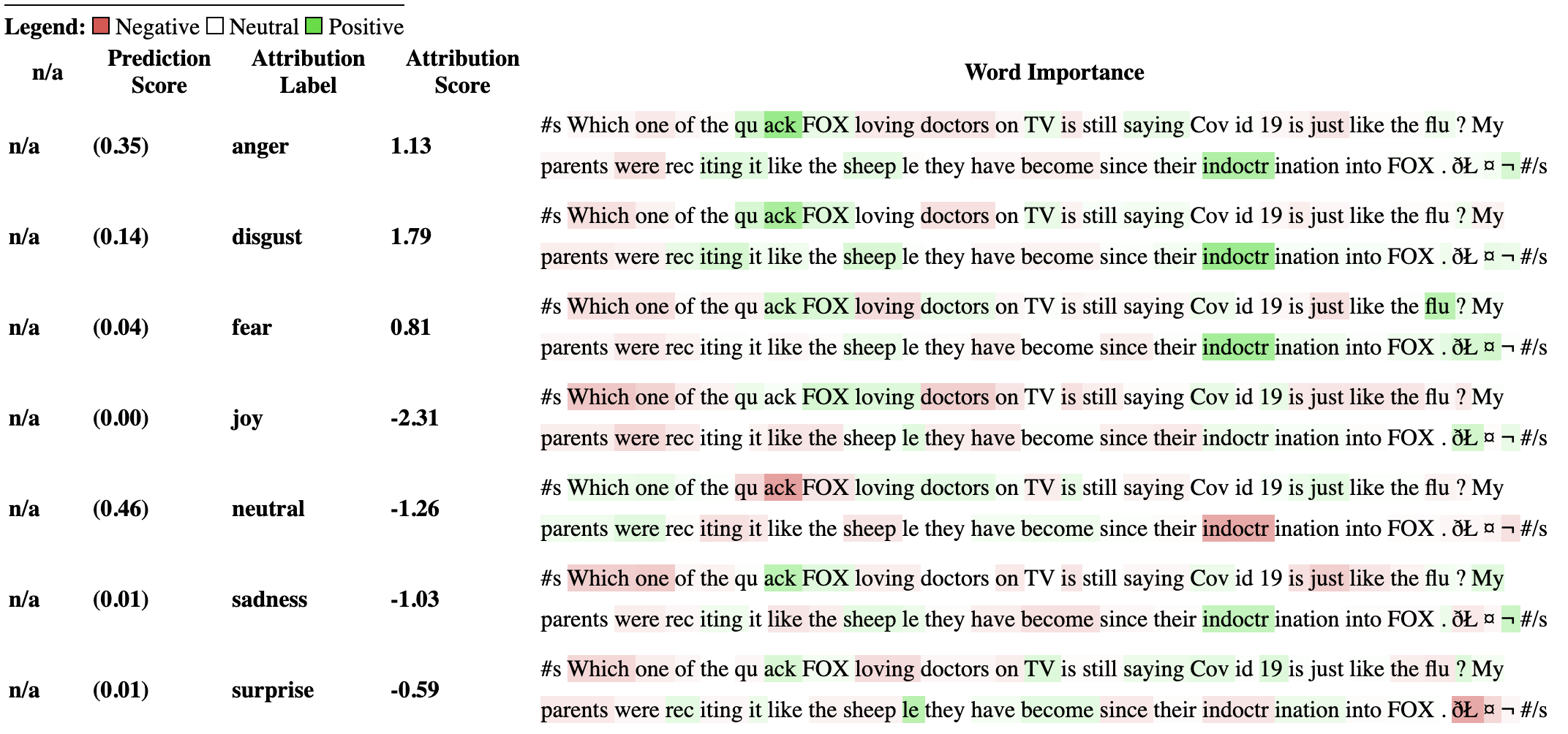}%
    \caption{A visualization of the output from our emotion encoder model analyzed with Captum, a popular model explainability tool. Each line shows possible classification. Green highlighting means a word has a positive effect on the classification score. Red highlighting means the word has a negative impact on the classification score. The intensity of the highlighting conveys the strength of the effect.}%
    \label{fig:Captum}%
\end{figure}
\begin{figure}%
    \centering
    \subfloat[\centering Train Data UMAP]{{\includegraphics[width=8cm, height=8cm]{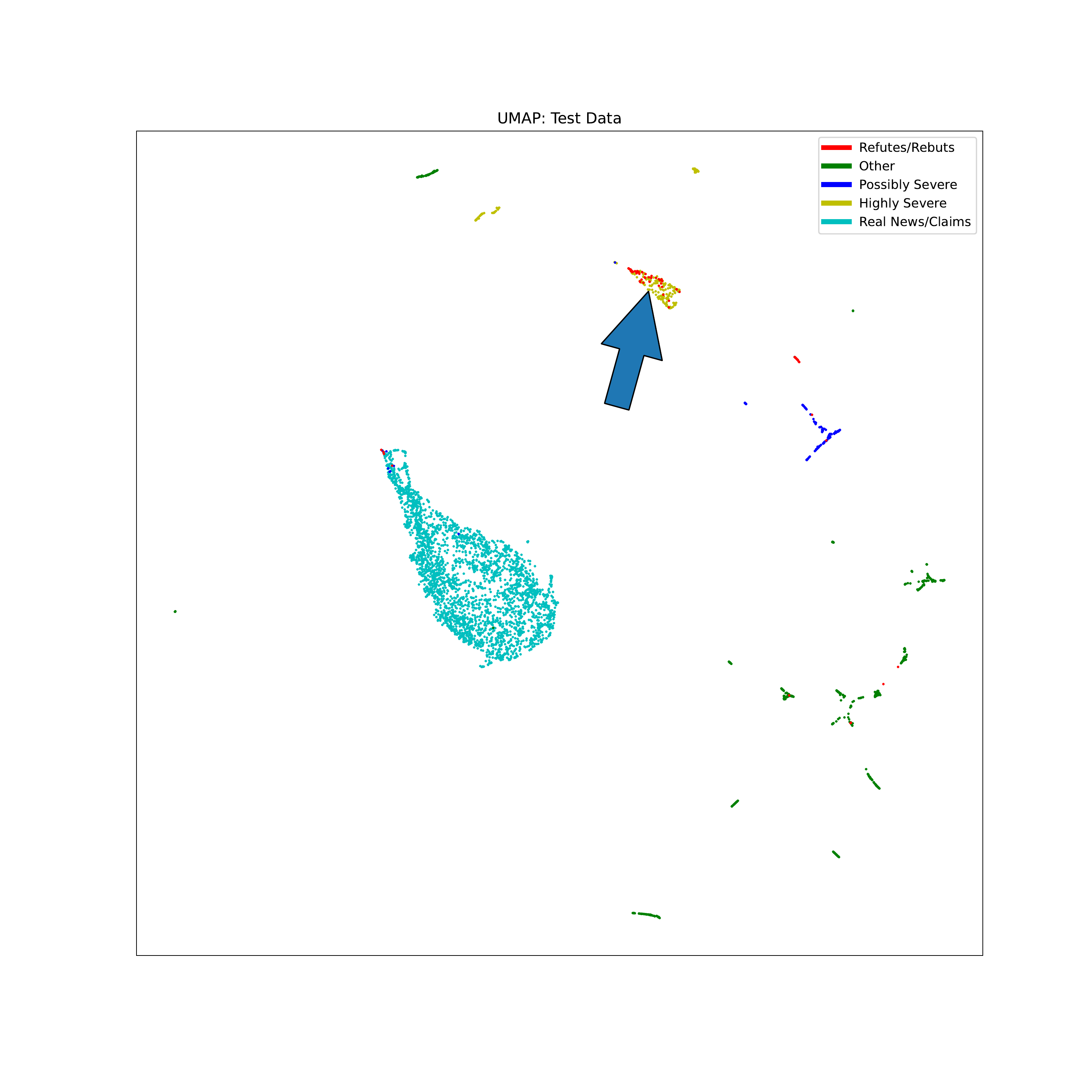} }}%
    \subfloat[\centering Test Data UMAP]{{\includegraphics[width=8cm, height=8cm]{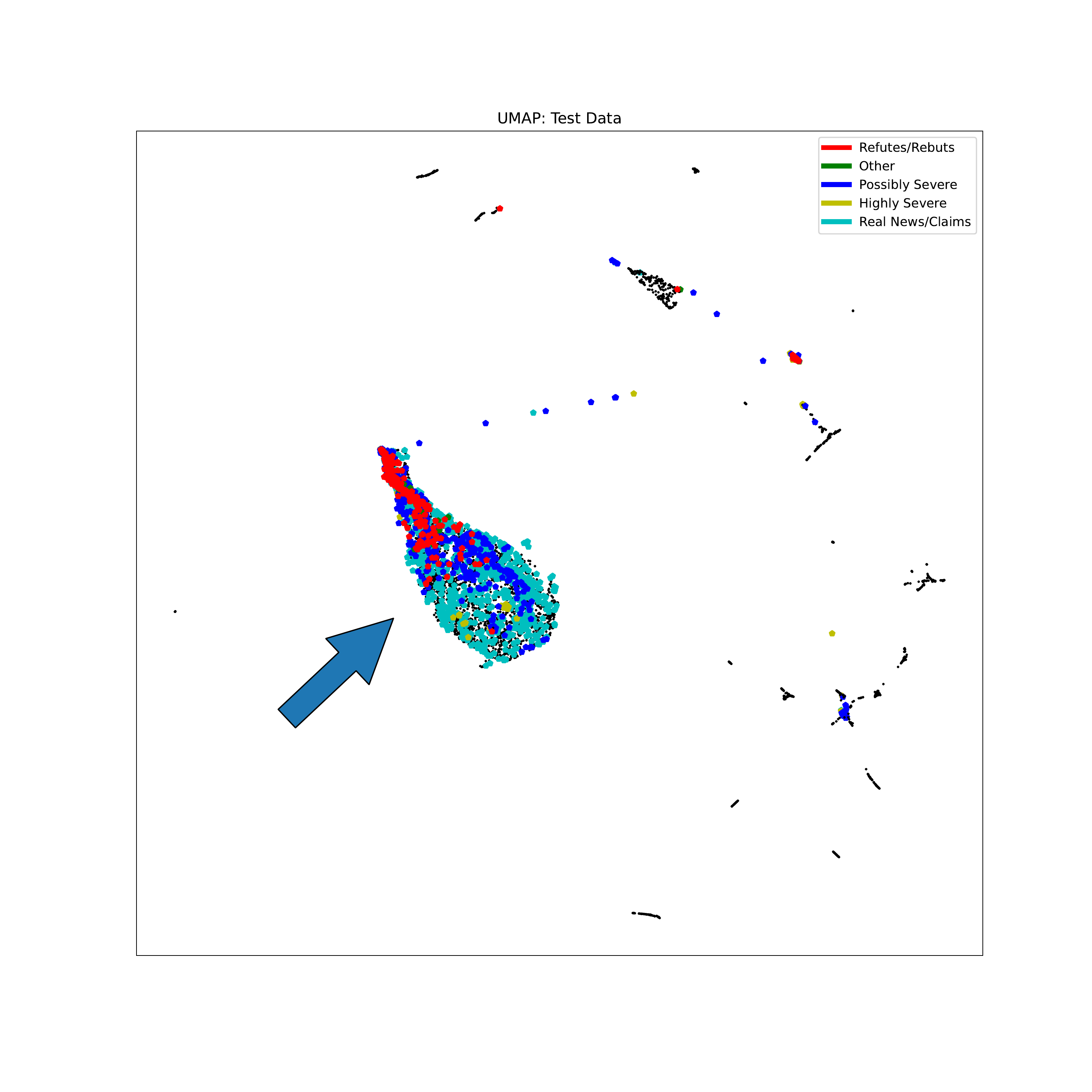} }}%
    \caption{UMAP figures for our training embeddings (a) and our train and test embeddings (b). The arrow in a) points to clustering of 'refutes/rebuts' and 'highly severe' misinformation labels together. The arrow in b) points to the model struggling classifying most samples as within the distribution of 'real news' labeled training tweets.}%
    \label{fig:UMAP}%
\end{figure}

We use Captum, a popular language model explainability tool, to analyze the outputs of our fine-tuned emotion encoder in our dual-encoder architecture \cite{kokhlikyan2020captum}. Figure \ref{fig:Captum} shows an example tweet and the tokens in the text that the model is paying attention to for each classification. In identifying emotions, like anger, disgust and fear, the model pays attention to charged words such as "quack", "sheep", and "indoctrination". These words have the inverse effect for predicting the neutral label. From this analysis, we observe that the model assigns importance to emotional words in a way that follows our intuition.

In addition to our Captum analysis, we extract train data embeddings from our COVID-19 HeRA fine-tuned SBERT classifier, and project these embeddings into a two-dimensional space (Figure \ref{fig:UMAP}) using UMAP (Uniform Manifold Approximation and Projection)\cite{mcinnes2018umap}. These UMAP plots visualize the direct representation that our model uses to classify given tweets. Thus, these plots offer deep insight into how the model classifies tweets. For example, in Figure \ref{fig:UMAP}, we observe two key trends. Firstly, our fine-tuned model clusters  "real-news", "other", and "possibly severe" tweets well. However, our model struggles to differentiate "highly severe" and "refutes/rebuts" labels, and clusters these two categories together. These two categories of tweets are often similarly structured, sharing the same informational content, and depend on tonal cues such as sarcasm for differentiation. Thus, it would logically follow, as evidenced by Figure \ref{fig:UMAP}, that the model would have difficulty distinguishing between these two classes. Furthermore, we also observe that the model has difficulty clustering our test set embeddings. As is evidenced in Figure \ref{fig:UMAP}, our model seems to be be clustering most test set tweets as "real-news" despite their ground truth label.

\subsection{Error Analysis and Limitations}
Our model errors can be attributed to four categories: distribution errors, COVID dataset quality errors, annotator baseline inconsistencies, and model architecture. 

\begin{figure}%
    \centering
    \subfloat{{\includegraphics[width=8cm, height=5cm]{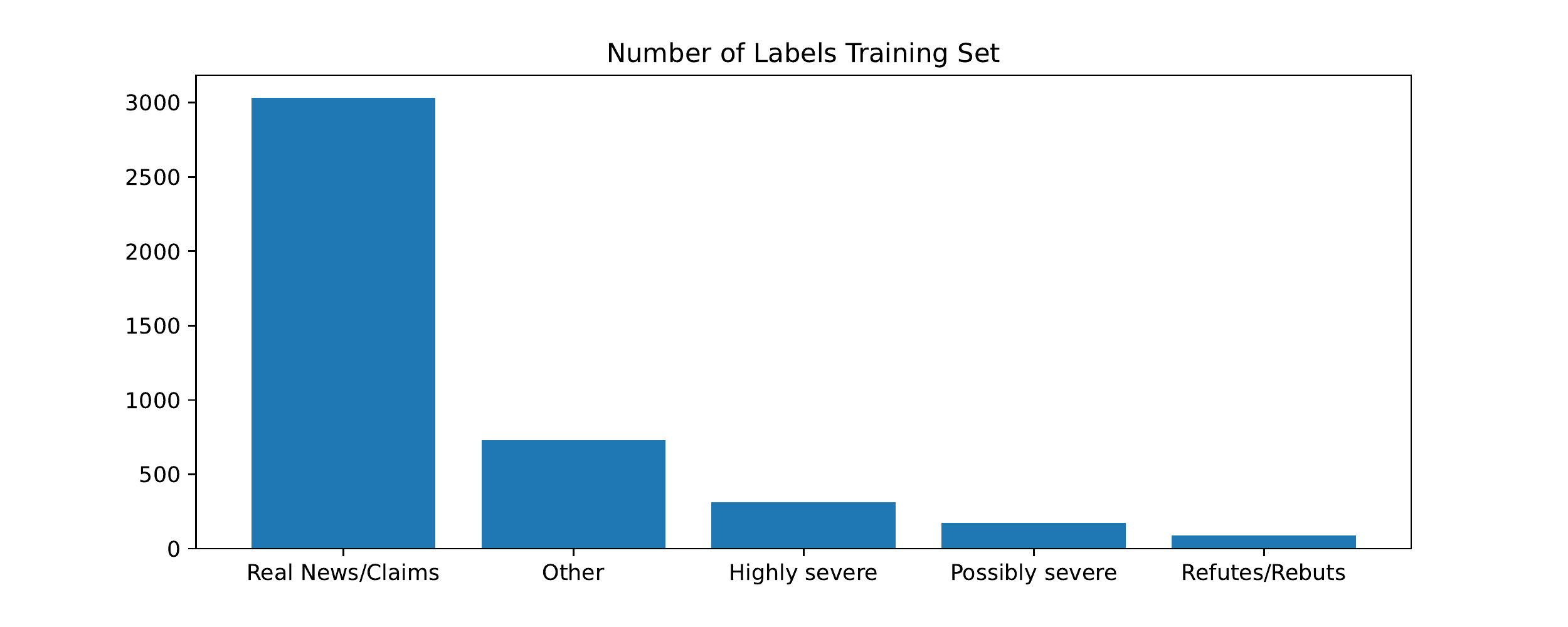} }}%
    \subfloat{{\includegraphics[width=8cm, height=5cm]{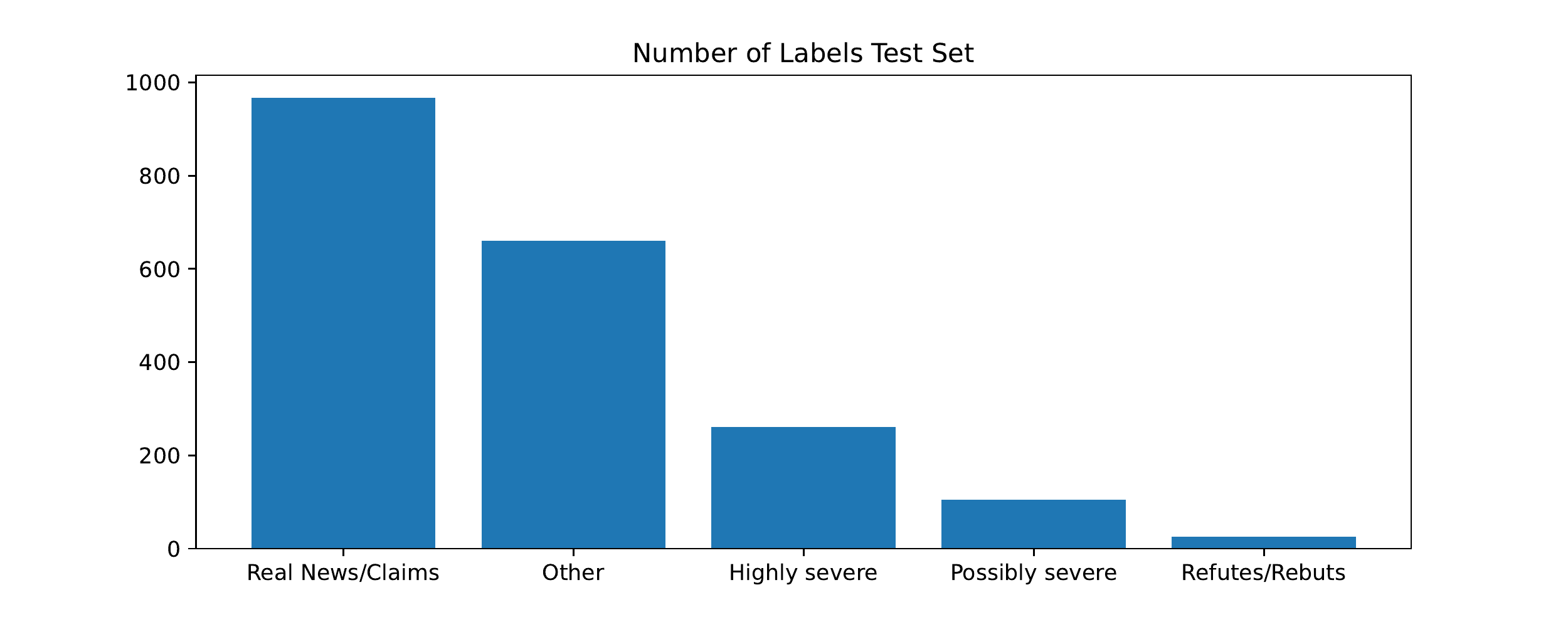} }}%
    \caption{Label distributions across training and test sets.}%
    \label{fig:label_distribution}%
\end{figure}

Firstly, a potentially significant source of error lies in the different distribution of COVID misinformation severity labels between the training and test sets. Figure \ref{fig:label_distribution} elucidates three key differences in distribution. Firstly, there are significantly more real news tweets as a proportion of data set size in the training data. Secondly, the proportion of "other" labeled tweets is far larger in the test set than training set. Finally, the number of "refutes/rebuts" class tweets is larger in the test dataset than training dataset\ref{dataset breakdown}. We believe that these differences in label distribution had adverse effects on our model. For example, during training, the model may have biased it's fit to "real news" tweets, which hurts performance on non real news labels. Furthermore, given that real news labeled tweets do not necessarily contain COVID-related information, the model may have learned to depend on features present in these real-news tweets which are irrelevant to predicting other categories.

\begin{table}[ht]
    \centering
    \begin{tabular}{|p{0.25\linewidth} | p{0.5\linewidth}|}
    \hline
    Error Category & Source text \\
    \hline
    Insufficient Information & “To all those people who said “COVID-19 is just like the flu”.,
    ... https://t.co/Vy4uGrJY8t” \textit{(Ground Truth: Refutes/Refuses)}
 \\
    \hline
    Real News Irrelevance & “Need to refresh your space? Here are some easy decor tips to get your Olympus Property home ready for spring! https://t.co/QcZTw0pWVG \#OlympusProperty \#OlympusProud \#ApartmentLiving \#HomeDecor \#LoveWhereYouLive \#ApartmentDecor \#Springtime \#YourHomeOurPassion https://t.co/wX8b4V4P4f” \textit{(Ground Truth: Real News)} \\
    \hline
    Label Ambiguity & @seanhannity How many stories did you do on the Coronavirus Hoax?
Maybe you should give TRUTH the Same Courtesy. \textit{(Ground Truth: Refutes/Refuses)}
\\
   \hline
    \end{tabular}
    \caption{\label{error categories} Example tweets for COVID-HeRA low quality annotation error categories}

\end{table}

The next significant source of errors lies in low quality annotations in the COVID-HeRA dataset. These low quality annotations can be separated into three categories, and examples shown in Table \ref{error categories}. Firstly, given that many tweets in our dataset were "retweets", there are many tweets that do not contain sufficient information to make a conclusive misinformation label prediction. For example, Table \ref{error categories} presents a tweet which is labeled as refuting and rebutting COVID misinformation, however this rebuttal is dependent on the link attached. Thus, without access to said link, the tweet reads as if it could also be an example of COVID misinformation. This lack of sufficient information to label tweets may cause our model to incorrectly associate features between classes. We also found many irrelevant tweets in our "real news" label (Table \ref{error categories}). These irrelevant tweets may cause our model to learn features completely irrelevant to COVID-19 as a workaround to label real-news tweets. This workaround defeats the purpose of a COVID-19 misinformation model, limiting the effectiveness of our model on unseen data. Finally, we detected many tweets with label ambiguity (Table \ref{error categories}). The misinformation level of these tweets is difficult to interpret even upon further tweet analysis. Often, these tweets are sarcastic or utilize higher level linguistic cues such as tone to inform label correctness. We suspect that the linguistic difficulty and ambiguity that these tweets contain presented a challenge to our model's learning. For example, as was shown in Figure \ref{fig:UMAP}, our model interprets tweets with "highly severe" and "refute/rebuts" labels similarly, which may owe to higher level linguistic cues such as sarcasm. These "refute/rebuts" tweets may be offering misinformation under a sarcastic tone, hence the difficulty distinguishing between labels.

Annotator label inconsistencies in our emotion dataset may have been another factor in causing model error. For example, in categories other than "neutral", there was often disagreement between the three annotators, resulting in potentially mislabeled labels. Furthermore, Ekman's 6 emotions, the emotion labels that we used, are useful for classification but do not encompass the nuance of language. They negate tonal shifts such as sarcasm, and do not capture ambiguous sub-emotions (loathing vs fear or disgust). These emotion label inconsistencies may have biased our emotion encoder and limited the effectiveness of this module on our COVID classification task.

There are also  model architectural limitations and errors. Primarily, there is a discongruence between the training data for SBERT, our encoder, and the COVID-HeRA dataset, the fine-tuned task. SBERT was trained on data sources such as Wikipedia, Stack Overflow, and Wiki-Answers. Thus, it was trained on data which contains full sentences, is semantically cohesive, and contains a relatively monotonic per-token signal, i.e the importance of words is more distributed across sentences. On the other hand, our model was fine-tuned on tweets. Tweets contain machine unreadable tokens such as emojis, contain incomplete sentences, and have a diverse and non-monotonic per-token signal, i.e one word is more important than all others in a tweet. 

Finally, we anticipate two more limitations to our work. Firstly, our emotion encoder utilized a small subset of data compared to our COVID misinformation classifier. This emotion encoder was fine-tuned on 100 tweets, which represents a small fraction of our total dataset size. Thus, we anticipate errors in emotion labels, which may have reduced the impact of adding an emotion classifier on top of a COVID-19 classifier. Second, due to the ever-changing progression of COVID-19, the categories of COVID misinformation adopted in the COVID-HeRA dataset may be obsolete. For instance, multiple U.S. government agencies have concluded that COVID-19 most likely leaked from a controlled laboratory environment \cite{Barnes_2023}. This "Lab Leak" theory of COVID-19's origin was originally thought to be misinformation, and the COVID-HeRA dataset treats supportive statements of it as such. For instance, the tweet "Look at the date of the article. China should be held accountable! \#coronavirus Lab-Made Coronavirus Triggers Debate | The Scientist Magazine®" is labeled as Other (Misinformation). Labels for tweets like this may need to be corrected for future research into identifying COVID-19 misinformation on Twitter.

\subsection{Future Work}

We believe our model architecture could be used to detect misinformation in other subject areas. Recent elections across multiple US states and countries have been the subject of conspiracy theories of vote tampering and fraud\cite{bieber_2022}. We also see promise in using our dual semantic / emotion encoding architecture outside of the Twitter domain. Rozado et al. showed in a longitudinal study that frequency of emotional text in news headlines has increased in recent years \cite{rozado-emotion-news}. It follows that encoding emotion in addition to content may help determine if news headlines contain misinformation.

\subsection{Conclusion}
Our study shows that encoding emotion in addition to content may improve the performance of standard transformer-based models in detecting the severity of misinformation on Twitter. Our novel architecture showed improvements in accuracy, recall and F1-score relative to the baseline. We also created a new hand-annotated dataset to fine tuning our emotion encoder model, and used it to add emotion encodings to a broader set of tweets from the COVID-HeRA dataset. Our analysis also showed that distinguishing highly emotional sarcastic comments from mis-informative ones is a difficult task for our models. Further work is required to understand the differences between those emotionally charged tweets that refute or rebut misinformation and those that promote it. Understanding the complex emotional tones of online communication seems to be a promising avenue for improving language model tasks that involve social media communication, where incidences of irony and sarcasm are much higher than the general corpus of text that language models are trained on.

\bibliography{bib}

\end{document}